\documentclass[11pt]{article}

\usepackage[preprint]{acl}

\usepackage{times}
\usepackage{latexsym}

\usepackage[T1]{fontenc}

\usepackage[utf8]{inputenc}

\usepackage{microtype}

\usepackage{inconsolata}

\usepackage{graphicx}
\usepackage{subcaption}
\usepackage{amsmath}
\usepackage{amssymb}
\usepackage{makecell}
\usepackage[ruled, vlined, linesnumbered]{algorithm2e}\DontPrintSemicolon 
\usepackage{comment}
\usepackage{enumitem}
\usepackage{adjustbox}
\usepackage{arydshln}

\title{AdaSD: Adaptive Speculative Decoding for Efficient Language Model Inference}

\author{
  \textbf{Kuan-Wei Lu\textsuperscript{1}},
  \textbf{Ding-Yong Hong\textsuperscript{1}},
  \textbf{Pangfeng Liu\textsuperscript{2}},
  \textbf{Jan-Jan Wu\textsuperscript{1}}
\\
\\
  \textsuperscript{1}Institute of Information Science, Academia Sinica
\\
  \textsuperscript{2}Department of Computer Science and Information Engineering, National Taiwan University
\\
  \small{
  \textbf{Correspondence:}
  \href{mailto:andylu6046@iis.sinica.edu.tw}{andylu6046@iis.sinica.edu.tw};
  \href{mailto:dyhong@iis.sinica.edu.tw}{dyhong@iis.sinica.edu.tw};
  \href{mailto:pangfeng@csie.ntu.edu.tw}{pangfeng@csie.ntu.edu.tw};
  \href{mailto:wuj@iis.sinica.edu.tw}{wuj@iis.sinica.edu.tw}
  }
}

\begin{document}
\maketitle

\begin{abstract}
Large language models (LLMs) have achieved remarkable performance across a wide range of tasks, but their increasing parameter sizes significantly slow down inference.
Speculative decoding mitigates this issue by leveraging a smaller draft model to predict candidate tokens, which are then verified by a larger target model.
However, existing approaches often require additional training, extensive hyperparameter tuning, or prior analysis of models and tasks before deployment.
In this paper, we propose Adaptive Speculative Decoding (AdaSD), a hyperparameter-free decoding scheme that dynamically adjusts generation length and acceptance criteria during inference.
AdaSD introduces two adaptive components: one to determine when to stop candidate token generation and the other to decide token acceptance, updated in real time based on token entropy and Jensen-Shannon distance.
This approach eliminates the need for pre-analysis or fine-tuning and is compatible with off-the-shelf models.
Experiments on benchmark datasets demonstrate that AdaSD achieves up to 1.46$\times$ speedup over vanilla speculative decoding while limiting accuracy degradation to under 1.8\%, making it a practical solution for efficient and adaptive LLM inference.
\end{abstract}

\section{Introduction}

Large language models (LLMs) have rapidly become integral to a wide range of applications, including natural language understanding, content generation, and code synthesis.
This progress is driven by the advances of the Transformer architecture~\cite{vaswani2017attention}.
Representative models such as T5~\cite{raffel2020exploring}, GPT~\cite{brown2020language}, Gopher~\cite{rae2022scaling}, and PaLM~\cite{chowdhery2023palm} have successively advanced the state of the art, demonstrating that scaling up model parameters can substantially improve performance across many real-world tasks.

Despite these remarkable performance gains, increasing model size also introduces substantial inference overhead.
For example, Meta's flagship model Llama 3~\cite{grattafiori2024llama3} achieves 87.3\% accuracy on the MMLU general knowledge benchmark with its 405 billion parameters.
However, at this massive scale, inference becomes memory-bound, as every forward pass requires loading the entire set of model weights from memory, which significantly slows down computation and poses challenges for practical deployment.

Speculative decoding~\cite{leviathan2023fast} is a technique to alleviate the memory-bound bottleneck by maximizing the utilization of each weight load during inference.
It uses a smaller and faster, but less accurate, \emph{draft model} to generate a sequence of candidate tokens, which are then verified \emph{in parallel} by the original large \emph{target model} to ensure correctness.
By offloading token generation to the draft model and verifying multiple tokens simultaneously, speculative decoding significantly improves inference throughput while preserving accuracy, effectively reducing the latency imposed by repeatedly loading the full set of model weights.

Building upon this foundation, several methods have been proposed to improve speculative decoding, such as fine-tuning the draft model to better align with the target model's distribution~\cite{zhou2024distillspec, li2024eagle, li2024eagle2}, dynamic mechanisms to determine the optimal length of the candidate token sequences~\cite{mamou2024dynamic, huang2025specdec, liu2025pearl}, and relaxing the acceptance criteria during target model verification~\cite{kim2023speculative, holsman2025fuzzy}.

However, these methods have limitations.
First, aligning the draft model often requires additional training, adding computational cost.
Second, relaxing acceptance criteria introduces trade-offs between accuracy and efficiency, and requires extra efforts to determine the tolerable levels.
Third, many approaches rely heavily on extensive hyperparameter tuning, necessitating iterative configuration searches to achieve satisfactory performance.

To address the aforementioned limitations, we propose \emph{Adaptive Speculative Decoding} (AdaSD), a hyperparameter-free approach to enhance LLM inference efficiency.
AdaSD consists of two components: one determines the termination of candidate token generation, and the other defines acceptance criteria for verification.
Specifically, we employ entropy and Jensen-Shannon distance as principled metrics to set these components separately.
Furthermore, both metrics are dynamically adjusted during inference based on information from previously generated tokens, eliminating the need for manual configuration or model-specific tuning.

The contributions of this paper are as follows:
\begin{itemize}[nosep]
    \item We propose AdaSD, an adaptive speculative decoding method that jointly adjusts candidate generation length and acceptance criteria to balance inference speed and accuracy.
    \item We introduce a Bayesian verification mechanism that estimates the acceptance probability of candidate tokens, providing an adaptive alternative to fixed verification rules.
    \item Our method is compatible with off-the-shelf models, requiring neither model architecture change nor additional training.
    Moreover, we eliminate the need for hyperparameter tuning.
    \item Experimental results indicate that AdaSD achieves up to 1.46$\times$ speedup over standard speculative decoding, while limiting accuracy degradation to less than 1.8\%.
\end{itemize}

\section{Background}
\subsection{Language Model Acceleration}
\subsubsection{Auto-Regressive Model}
An auto-regressive model employs a decoder-based Transformer architecture to perform text generation.
It produces each output token conditioned on the preceding tokens.
The newly generated token is then appended to the input sequence, serving as the context for the next step of generation.

Let $p$ denote the probability distribution of the auto-regressive model.
At each time step $t$, the next token $x_t$ is drawn from the conditional probability distribution over the preceding token sequence $x_{<t}$, defined as $x_t \sim p(x \mid x_{<t})$.
The token $x_t$ can be selected \emph{deterministically} by choosing the most probable token: $x_t = \arg\max_{x} p(x \mid x_{<t})$.
Alternatively, $x_t$ can be chosen \emph{stochastically} by sampling from the adjusted distribution, using techniques such as temperature scaling, top-$k$ sampling, or top-$p$ sampling.
This iterative process continues until the model produces a special end-of-sequence (EOS) token, which terminates generation.

\subsubsection{Speculative Decoding}

Speculative decoding is designed to address the high inference time of auto-regressive generation in the large language model, denoted as the \emph{target model} $M_p$.
To mitigate this, a smaller and faster model, referred to as the \emph{draft model} $M_q$, is introduced to approximate the target model's output distribution.

Each iteration of speculative decoding consists of two steps: \emph{generation} and \emph{verification}.
In the generation step, the draft model proposes candidate tokens by sequentially producing a number of tokens up to a limit, which may be fixed or adaptively adjusted.
In the verification step, the target model evaluates these candidate tokens in parallel by expanding them into a batch and accepting the longest valid prefix to ensure correctness.
By combining fast speculative generation with efficient verification, speculative decoding lowers inference latency without compromising output quality.

\subsubsection{Sampling Strategy}

To determine which tokens are accepted during the verification step of speculative decoding, a straightforward deterministic strategy is \emph{greedy decoding}.
In this approach, both the draft model $M_q$ and the target model $M_p$ always select the token with the highest probability:
\begin{gather*}
    x_t^q = \arg\max_x M_q(x \mid x_{<t}), \\
    x_t^p = \arg\max_x M_p(x \mid x_{<t}),
\end{gather*}
where $x_t^q$ and $x_t^p$ are the tokens chosen by the draft and target models at step $t$, respectively.
Greedy decoding imposes a strict acceptance rule: the candidate token $x_t^q$ is accepted only if it exactly matches $x_t^p$.
The verification process determines the longest valid prefix by scanning the evaluation results in parallel until the first mismatch occurs.

Although greedy decoding is simple, it is overly restrictive and limits diversity in text generation.
To address this limitation, prior work~\cite{leviathan2023fast, chen2023accelerating} introduces \emph{speculative sampling}, which relaxes the strict matching requirement by allowing tokens from the draft model to be probabilistically accepted by the target model.
Speculative sampling preserves the target distribution by jointly considering the distributions of both models.
A candidate token $\tilde{x_t}$ is first sampled from the draft model distribution and then accepted by the target model with probability $\alpha$:
\begin{gather*}
    \tilde{x_t} \sim M_q(x \mid x_{<t}), \\
    \alpha = \min\left(1, \frac{M_p(\tilde{x_t} \mid x_{<t})}{M_q(\tilde{x_t} \mid x_{<t})}\right).
\end{gather*}
If the candidate token is rejected, the target model re-samples a new token from an adjusted distribution that excludes $\tilde{x_t}$.

\subsection{Jensen-Shannon Divergence and Distance}

Entropy and cross-entropy provide fundamental measures for characterizing uncertainty and distribution mismatch.
Building on these concepts, Kullback-Leibler (KL) divergence offers a principled way to quantify the difference between two probability distributions.
For completeness, the formal definitions of entropy, cross-entropy, and KL divergence are provided in Appendix~\ref{sec:information_theoretic_background}.

However, KL divergence is asymmetric and unbounded, making it less desirable for directly comparing two probability distributions.
Therefore, Jensen-Shannon (JS) divergence is often adopted as a symmetric and bounded alternative derived from KL divergence.
JS divergence is defined as:
\begin{equation*}
    \begin{split}
        D_\mathrm{JS}(p \parallel q) &= \frac{1}{2} D_\mathrm{KL}(p \parallel m) + \frac{1}{2} D_\mathrm{KL}(q \parallel m) \\
                                     &= H(m) - \frac{1}{2} (H(p) + H(q)),
    \end{split}
\end{equation*}
where $m = \frac{1}{2}(p+q)$ is the mixture distribution of the draft model $q$ and the target model $p$, $H(\cdot)$ denotes entropy, and $D_\mathrm{KL}(\cdot \parallel m)$ denotes the KL divergence with respect to $m$.
Unlike KL divergence, JS divergence is symmetric in $p$ and $q$, and is bounded in $[0, 1]$ under base-2 logarithm.
These properties make it a more stable and interpretable metric for quantifying the dissimilarity between probability distributions.

To provide a more intuitive measure of the difference between distributions, we consider the Jensen-Shannon (JS) distance, defined as the square root of the JS divergence:
\begin{equation*}
    d_\mathrm{JS}(p \parallel q) = \sqrt{D_\mathrm{JS}(p \parallel q)}.
\end{equation*}
Since the JS distance satisfies the properties of a metric~\cite{endres2003a}, it provides a well-defined notion of distance between probability distributions within the metric space it induces.

\section{Related Work}
Several studies enhance speculative decoding by modifying model architectures.
Medusa~\cite{cai2024medusa} retains a single model but equips it with multiple decoding heads, enabling parallel prediction of consecutive future tokens at each step.
EAGLE~\cite{li2024eagle, li2024eagle2} reconstructs draft transformer layers via additional training and employs dynamic tree search to align with the target distribution.
Both approaches, however, entail significant model adaptation and require extra training before yielding performance gains.

To mitigate the memory and computational overhead associated with deploying a separate auxiliary model, some works focus on self-speculative decoding.
This paradigm leverages the internal structure of the target model itself to act as the draft model, while the full model is subsequently used for verification.
Draft \& Verify~\cite{zhang2024draft} selectively skips intermediate layers during the draft step based on an offline Bayesian optimization search.
EESD~\cite{liu2024speculative} constructs its draft model by introducing a trainable early-exiting layer immediately following the first few layers of the target model.
SWIFT~\cite{xia2025swift} dynamically optimizes the set of skipped layers on-the-fly during inference without additional training.
However, although self-speculative approaches eliminate the extra memory footprint, their speedup remains inherently bounded.
Omitting an excessive number of layers to accelerate drafting inevitably degrades the prediction quality and lowers the acceptance rate.

Another line of research focuses on adjusting the candidate token generation length in draft models.
DISCO~\cite{mamou2024dynamic} employs a two-layer feedforward network (FFN) classifier to adaptively set a termination threshold, stopping generation once a candidate token's probability falls below the threshold.
Hugging Face's Assisted Generation~\cite{gante2023assisted} extends DISCO to an unsupervised variant that preserves the same decision mechanism but replaces the FFN classifier with a receiver operating characteristic (ROC) curve.
AdaEDL~\cite{agrawal2024adaedl} introduces a training-free early stopping criterion that halts the drafting process by approximating a lower bound on the expected token acceptance probability using the entropy of the draft model's logits.
SpecDec++~\cite{huang2025specdec} formulates the problem as a Markov decision process and introduces a trained head that predicts the acceptance probability of candidate tokens to decide when to stop.
PEARL~\cite{liu2025pearl} simulates dynamic draft lengths through two strategies: it first verifies whether a newly generated token is accepted by the target model, and upon success, allows the draft model to continue generating tokens in parallel with verification, creating the effect of dynamic length adjustment.
HSDDW~\cite{syu2025hierarchical} proposes a three-layer hierarchical structure of progressively larger models.
The smallest model applies a threshold-based mechanism to control generation or forward tokens to larger models in the hierarchy for verification.
C2T~\cite{huo2026c2t} leverages a trainable lightweight classifier based on joint probability, entropy, and depth to dynamically pre-prune token trees and facilitate early stopping during inference.
Although effective, these approaches primarily focus on the generation process of draft models, and leave the target model's role in verification unexplored.

A number of works define acceptance criteria based on the divergence between draft and target distributions.
BiLD~\cite{kim2023speculative} uses a fixed cross-entropy threshold to determine whether to accept draft tokens, while FSD~\cite{holsman2025fuzzy} utilizes a fixed JS divergence threshold, accepting tokens when the divergence between draft and target models is sufficiently small.
However, both methods require model- or task-specific information in advance for deciding the optimal thresholds, which limits their general applicability.

The summary and comparison of related methods are provided in Appendix~\ref{sec:method_comparison}.
\section{Adaptive Speculative Decoding}
In this section, we first provide an overview of AdaSD.
We then present the empirical study, whose insights motivate this work.
Finally, we describe the proposed method in detail.

\subsection{Overview}
We propose Adaptive Speculative Decoding (AdaSD), a hyperparameter-free scheme for efficient LLM inference.
AdaSD differs from prior methods in that it requires neither additional training nor any model- or task-specific information in advance.
Moreover, it optimizes the decoding decisions by taking both the draft and target models into account.
In particular, it eliminates the burden of hyperparameter tuning, which often requires exhaustive search and manual effort.

\begin{figure*}[t!]
    \centering
    \includegraphics[width=.8\linewidth]{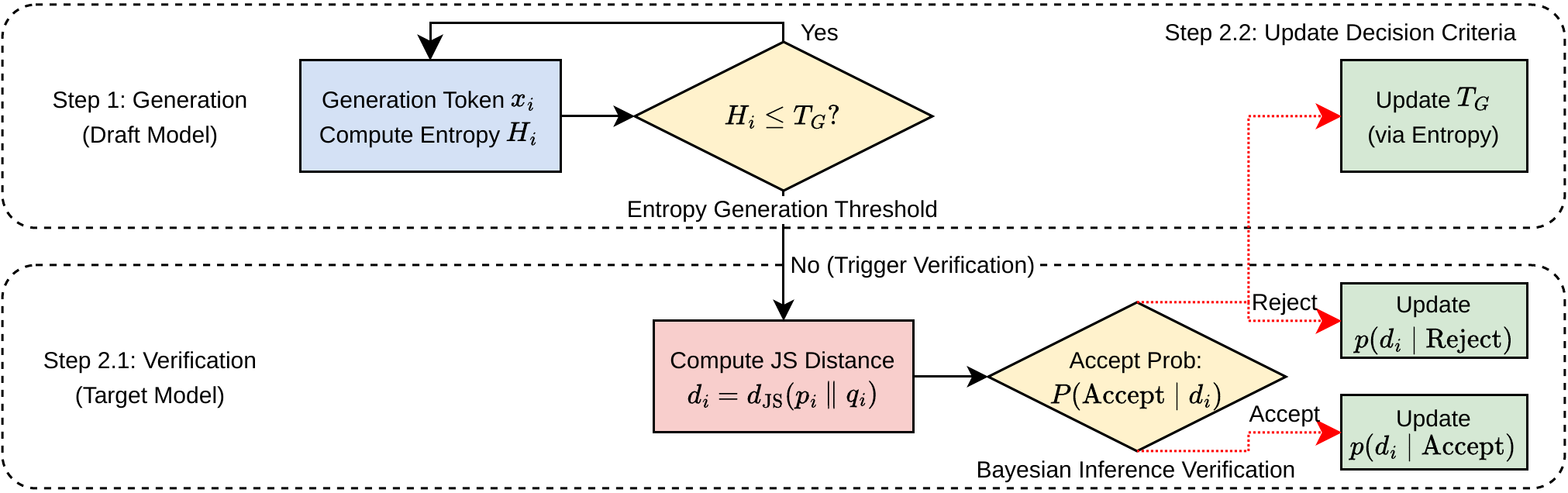}
    \caption{The decoding process of AdaSD.}
    \label{fig:adasd}
    \vspace{-2mm}
\end{figure*}

AdaSD consists of two steps.
The \emph{generation} step (Section~\ref{sec:generation}) determines when the draft model should stop generating candidate tokens and hand them off to the target model.
The \emph{verification} step (Section~\ref{sec:verification}) regulates the acceptable level of dissimilarity between the probability distributions of the draft and target models.
The workflow of AdaSD is illustrated in Figure~\ref{fig:adasd}.

To further improve efficiency, AdaSD introduces a heuristic feedback mechanism that adaptively adjusts decision rules during inference.
Instead of relying on pre-analysis of models or tasks, AdaSD dynamically leverages statistics from previously generated tokens to compute entropy and JS distance to update decision rules on the fly.
The algorithm of our method is provided in Appendix~\ref{sec:algorithm_of_adasd}.

\subsection{Empirical Study}
\label{sec:empirical_study}
We begin by considering the generation step of the draft model.
Since entropy quantifies the uncertainty associated with a \emph{single} random variable, we use it solely within the draft model to estimate the uncertainty of its generated candidate tokens, independent of the target model.
This uncertainty measurement is commonly employed in early-exit networks~\cite{agrawal2024adaedl}.

For the verification, we propose a joint optimization approach that simultaneously considers both the draft and target models.
The main challenge lies in selecting an appropriate indicator that captures the correlation between the two models and determines the acceptance probability of tokens.
At the same time, the indicator must be {\em bounded} to ensure that the resulting acceptance probabilities remain within a well-defined range, facilitating stable analysis and implementation.

Cross-entropy and KL divergence are unsuitable in this context because, when the distributions $p$ and $q$ differ substantially, their values can diverge to infinity, making the threshold unmanageable.
To address this issue, we propose using JS distance.
While both JS divergence and JS distance are bounded, JS distance offers the additional advantage of defining a proper metric, providing a more stable and consistent signal for adaptive verification within a metric space.

\begin{figure}[t!]
    \centering
    \includegraphics[width=.9\linewidth]{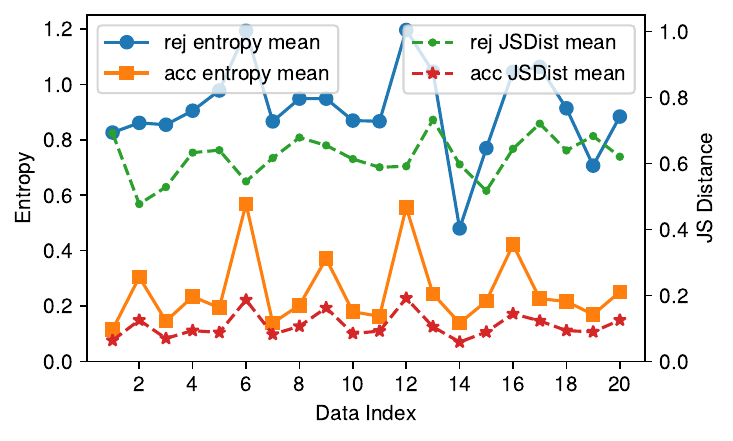}
    \caption{Mean entropy and Jensen-Shannon distance for accepted and rejected tokens, evaluated individually on 20 samples from Alpaca dataset, using Llama 3.1 8B and 70B as the draft and target models, respectively.}
    \label{fig:mean_entr_jsdist}
\end{figure}

To validate the effectiveness of these measures, we conduct an initial experiment analyzing entropy and JS distance for accepted and rejected tokens.
We sample 20 examples from the widely used instruction-tuning dataset Alpaca~\cite{taori2023alpaca}, and use Llama 3.1 8B as the draft model and Llama 3.1 70B as the target model.
For each example, we compute the mean entropy of the candidate token distributions produced by the draft model, as well as the mean JS distance between the draft and target models for both accepted and rejected tokens.
The results, illustrated in Figure~\ref{fig:mean_entr_jsdist}, show a clear separation.
The rejected tokens consistently exhibit higher entropy and JS distance than the accepted tokens, suggesting that both measures can serve as effective signals for predicting token acceptance and guiding the design of adaptive thresholds.

\begin{figure}[t!]
    \centering
    \includegraphics[width=.9\linewidth]{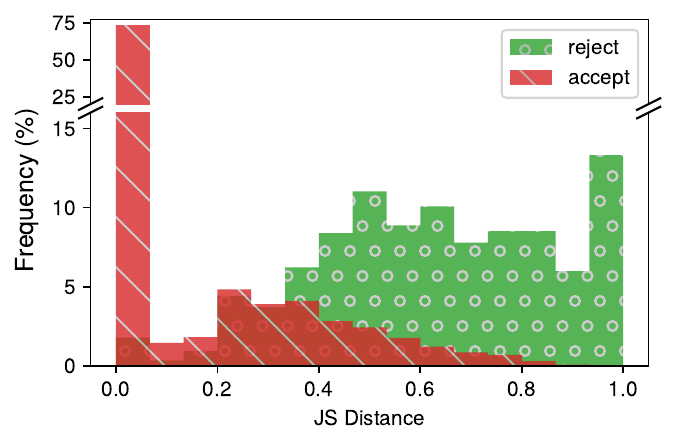}
    \caption{Jensen-Shannon distance distribution of accepted and rejected tokens.}
    \label{fig:distribution_jsdist}
\end{figure}

To further understand the correlation between models, we examine the distribution of JS distances across the sample data.
As shown in Figure~\ref{fig:distribution_jsdist}, a certain samples of rejected tokens exhibits JS distances close to 1, whereas such cases would produce extreme values that destabilize the threshold when using cross-entropy or KL divergence.
Furthermore, the JS distance distribution for accepted tokens shows two distinct clusters.
The first cluster (i.e., near zero) occurs when both models assign most of the probability mass to the \emph{same} token, typically in highly predictable contexts.
The second cluster appears when both models distribute probability more evenly across multiple options but ultimately select the same token, usually in contexts with weaker contextual correlation.
These results offer insights that JS distance effectively captures model agreement and measures uncertainty, allowing to provide robust thresholding for token verification in AdaSD.

\subsection{Generation}
\label{sec:generation}

The generation step in AdaSD employs an entropy-based threshold to control candidate token generation in the draft model.
If the entropy of a newly generated candidate token exceeds this threshold, indicating that the token is likely to be rejected, the generation process is terminated and control proceeds to the verification step.

The generation threshold is updated adaptively during inference.
Let $e_r$ denote the entropy of a candidate token rejected by the target model during the verification step.
We define $E_r$ as the set of all such entropy values collected since the start of the decoding process.
The generation threshold $T_G$ is then recalculated as the mean value of the elements in $E_r$:
\begin{equation*}
    T_G = \frac{1}{|E_r|}\sum_{e \in E_r} e.
\end{equation*}
This adaptive mechanism ensures that $T_G$ dynamically adjusts based on the draft model's historical uncertainty for rejected tokens.

\subsection{Verification}
\label{sec:verification}

At each verification step, we measure the discrepancy between the draft and target models by computing the JS distances between their output probability distributions for the candidate tokens.
The JS distance provides a symmetric and bounded measure of divergence and serves as the primary signal for regulating token acceptance during verification.

Instead of filtering candidate tokens using a fixed threshold as in previous approaches~\cite{kim2023speculative, holsman2025fuzzy},
we model the {\em acceptance probability} of a candidate token as a function of the JS distance.
Intuitively, smaller distances indicate stronger agreement between the draft and target models and should therefore correspond to higher acceptance probabilities.

To capture this relationship, we employ {\em Bayesian inference}.
Given the JS distance $d$ of a candidate token, the probability of accepting the token is computed from the relative likelihood under two density functions:
\begin{equation*}
    P(A \mid d) = \frac{p(d \mid A) P(A)}{p(d \mid A) P(A) + p(d \mid R) P(R)},
\end{equation*}
where $A$ and $R$ respectively denote the acceptance and rejection events, $P(A)$ and $P(R)$ represent the prior probabilities, while ${p(d \mid A)}$ and ${p(d \mid R)}$ are the conditional probability density functions of JS distances.
To estimate these densities, we collect the JS distances of verified tokens from the beginning of decoding and group them according to their verification outcome (accepted or rejected).
We then construct the empirical densities ${p(d \mid A)}$ and ${p(d \mid R)}$ using kernel density estimation (KDE)~\cite{silverman1986density}.

Note that we exclude tokens whose JS distances are exactly 0 or 1 when computing the KDEs.
This design provides three advantages.
First, distances of 0 and 1 correspond to cases of absolute agreement and disagreement between the draft and target models.
Excluding these cases allows the model to focus on estimating acceptance probabilities for the intermediate distances where the decision is uncertain.
Second, computing the KDE incurs a cost proportional to the number of collected samples.
As shown in Figure~\ref{fig:distribution_jsdist}, a significant portion of tokens have JS distances equal to 0 or 1.
Removing these tokens therefore reduces the number of samples involved in density estimation and significantly lowers the verification overhead without affecting the decision quality.
Third, KDE estimation may suffer from boundary leakage, where probability density spills outside the $[0, 1]$ range.
Excluding boundary cases therefore improves the accuracy of the density estimation.

\begin{figure}[t!]
    \centering
    \includegraphics[width=\linewidth]{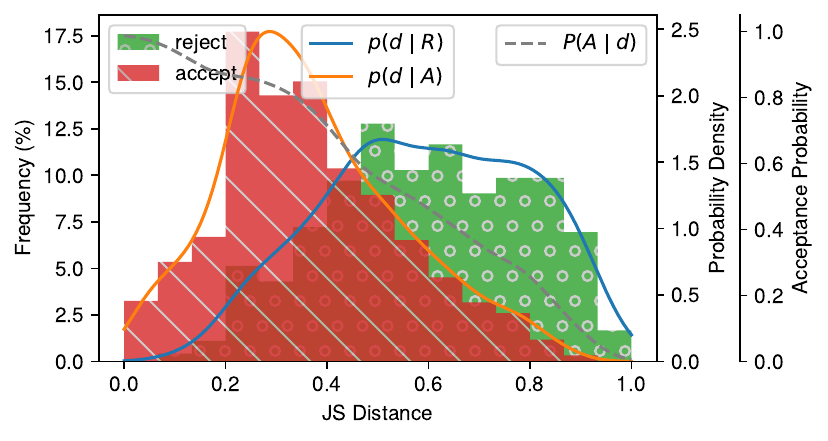}
    \caption{
        JS distance distribution for accepted and rejected tokens ($d = 0$ and $1$ filtered).
        We estimate probability densities with KDE and compute the acceptance probability as a function of JS distance.
    }
    \label{fig:filtered_distribution}
\end{figure}

Figure~\ref{fig:filtered_distribution} illustrates the acceptance probability as a function of the JS distance for the samples in Figure~\ref{fig:distribution_jsdist}, excluding the distances of 0 and 1.
The orange and blue curves correspond to the KDEs for accepted and rejected tokens, respectively.
The dashed curve represents the resulting acceptance probability derived from these density estimates, confirming our intuition that smaller distances exhibit higher acceptance probabilities.

After the target model makes the verification decision, JS distances of candidate tokens are added to the corresponding group (accepted or rejected), and the density estimates are updated accordingly.
Through this online update process, the acceptance criteria gradually adapts to the empirical distribution of verification outcomes during decoding.

\section{Experiments}
\subsection{Experimental Setup}
\paragraph{Benchmark.}
We select three benchmark datasets, each focusing on a different task.
GSM8K~\cite{cobbe2021training} contains grade school-level mathematical word problems.
HumanEval~\cite{chen2021evaluating} evaluates programming and code generation.
MMLU~\cite{hendrycks2021measuring} covers multiple-choice questions across a wide range of subjects.

\paragraph{Model.}
We select two representative model families, Llama 3~\cite{grattafiori2024llama3} and Qwen 2.5~\cite{yang2025qwen25}, 
and consider three target-draft model pairs: (Llama 3.1 70B, Llama 3.1 8B), (Llama 3.1 70B, Llama 3.2 1B), and (Qwen 2.5 72B, Qwen 2.5 7B).
These configurations enable a comprehensive evaluation of performance within the same model family across different scales and across model families of comparable sizes.

\paragraph{Baseline.}
We compare two speculative decoding methods that require no additional training and are agnostic to task-specific information:
(a) Vanilla speculative decoding~\cite{leviathan2023fast}, which uses a fixed candidate token length of five, and
(b) Hugging Face's built-in Assisted Generation~\cite{gante2023assisted}.
In the following experiments, we use \emph{Vanilla} and \emph{AssistedGen} to represent both methods, respectively.

\subsection{Performance Results}

Table~\ref{tab:main_result} presents the throughput and accuracy results for AdaSD and the baseline methods across all evaluated benchmarks, with more detailed data listed in Appendix~\ref{sec:additional_experimental_results}.
The performance of standalone draft and target models is included for reference, with efficiency measured as the speedup relative to the Vanilla baseline on four NVIDIA A6000 GPUs.
All evaluated methods use sampling, which may result in minor accuracy deviations from the original target model.
Detailed hardware and software configurations are provided in Appendix~\ref{sec:implementation_details}.

\begin{table*}[t!]
    \centering
    \adjustbox{max width=.8\linewidth}{
        \begin{tabular}{c|rcr|rcr|rcr}
    \hline
    & \multicolumn{3}{c|}{\textbf{GSM8K}} & \multicolumn{3}{c|}{\textbf{HumanEval}} & \multicolumn{3}{c}{\textbf{MMLU}} \\
    \hline
    \hline
    \multicolumn{10}{c}{\textbf{Llama 3.1 70B -- Llama 3.1 8B}} \\
    \hline
    & tks/sec & accuracy & speedup
    & tks/sec & accuracy & speedup
    & tks/sec & accuracy & speedup \\
    \hline
    Draft & 36.446 & 0.828 & 2.624 & 36.257 & 0.634 & 2.532 & 33.274 & 0.695 & 4.206 \\
    \hdashline
    Target & 4.842 & 0.939 & 0.349 & 4.838 & 0.750 & 0.338 & 4.376 & 0.836 & 0.553 \\
    Vanilla & 13.890 & 0.945 & 1.000 & 14.319 & 0.768 & 1.000 & 7.912 & 0.835 & 1.000 \\
    AssistedGen & 16.062 & 0.944 & 1.156 & 18.137 & 0.756 & \underline{1.267} & 7.970 & 0.835 & 1.007 \\
    \hdashline
    Gen-Only & 16.243 & 0.943 & \underline{1.169} & 18.017 & 0.762 & 1.258 & 8.023 & 0.831 & 1.014 \\
    Verify-Only & 14.227 & 0.931 & 1.024 & 14.425 & 0.750 & 1.007 & 8.247 & 0.836 & \underline{1.042} \\
    AdaSD & 17.039 & 0.943 & \textbf{1.227} & 18.348 & 0.799 & \textbf{1.281} & 8.477 & 0.834 & \textbf{1.071} \\
    \hline
    \hline
    \multicolumn{10}{c}{\textbf{Llama 3.1 70B -- Llama 3.2 1B}} \\
    \hline
    & tks/sec & accuracy & speedup
    & tks/sec & accuracy & speedup
    & tks/sec & accuracy & speedup \\
    \hline
    Draft & 99.919 & 0.375 & 6.080 & 100.683 & 0.348 & 5.764 & 95.141 & 0.453 & 10.878 \\
    \hdashline
    Target & 4.842 & 0.939 & 0.295 & 4.838 & 0.750 & 0.277 & 4.376 & 0.836 & 0.500 \\
    Vanilla & 16.434 & 0.939 & 1.000 & 17.467 & 0.756 & 1.000 & 8.746 & 0.838 & 1.000 \\
    AssistedGen & 19.892 & 0.936 & 1.210 & 23.265 & 0.774 & 1.332 & 8.920 & 0.833 & 1.020 \\
    \hdashline
    Gen-Only & 21.126 & 0.939 & \underline{1.286} & 24.842 & 0.768 & \underline{1.422} & 9.085 & 0.835 & 1.039 \\
    Verify-Only & 17.063 & 0.926 & 1.038 & 17.710 & 0.726 & 1.014 & 9.346 & 0.817 & \underline{1.069} \\
    AdaSD & 22.225 & 0.937 & \textbf{1.352} & 25.452 & 0.774 & \textbf{1.457} & 9.795 & 0.815 & \textbf{1.120} \\
    \hline
    \hline
    \multicolumn{10}{c}{\textbf{Qwen 2.5 72B -- Qwen 2.5 7B}} \\
    \hline
    & tks/sec & accuracy & speedup
    & tks/sec & accuracy & speedup
    & tks/sec & accuracy & speedup \\
    \hline
    Draft & 38.796 & 0.867 & 2.667 & 38.724 & 0.720 & 2.840 & 38.544 & 0.725 & 3.667 \\
    \hdashline
    Target & 4.728 & 0.913 & 0.325 & 4.717 & 0.787 & 0.346 & 4.690 & 0.845 & 0.446 \\
    Vanilla & 14.546 & 0.912 & 1.000 & 13.634 & 0.799 & 1.000 & 10.511 & 0.837 & 1.000 \\
    AssistedGen & 18.721 & 0.917 & \textbf{1.287} & 17.482 & 0.799 & \textbf{1.282} & 11.436 & 0.837 & \underline{1.088} \\
    \hdashline
    Gen-Only & 18.054 & 0.920 & 1.241 & 16.733 & 0.787 & 1.227 & 10.735 & 0.841 & 1.021 \\
    Verify-Only & 14.705 & 0.920 & 1.011 & 13.838 & 0.787 & 1.015 & 11.042 & 0.837 & 1.051 \\
    AdaSD & 18.289 & 0.915 & \underline{1.257} & 17.041 & 0.799 & \underline{1.250} & 11.470 & 0.839 & \textbf{1.091} \\
    \hline
\end{tabular}
    }
    \caption{
        Performance comparison of speculative decoding methods.
        Metrics include decoding speed (tks/sec), accuracy, and speedup over the Vanilla baseline.
        Gen-Only and Verify-Only represent AdaSD variants that use only the generation and verification component, respectively.
        The bold and underlined values indicate the highest and second-highest speedups.
    }
    \label{tab:main_result}
\end{table*}

Overall, AdaSD achieves the highest speedup among all methods on Llama models and is only slightly slower than AssistedGen on the Qwen model.
For GSM8K and HumanEval, AdaSD achieves 23-46\% higher throughput than Vanilla.
Compared to AssistedGen, AdaSD performs comparably or achieves further speedups depending on the model pairs (up to 14.2\% higher on GSM8K with Llama 3.1 70B--1B), while preserving accuracy within 1.8\% of the baseline.

A notable observation is the relatively modest speedup achieved by AdaSD on the MMLU dataset.
Both AdaSD and AssistedGen exhibit limited acceleration, with the highest improvement being only 12\%.
This is because MMLU is a multiple-choice benchmark, where the model generates only a small number of tokens corresponding to the selected answer.
The short sequence of these outputs inherently limits opportunities for acceleration.

We also observe that AdaSD achieves lower speedup than AssistedGen with Qwen 2.5.
To investigate this, we analyze the number of generated and accepted tokens, as well as the corresponding acceptance rates.
As Table~\ref{tab:qwen_analysis} shows, both the average lengths of generated and accepted tokens in AdaSD are longer than those in AssistedGen.
Although AssistedGen produces shorter sequence per speculative decoding step, it can still generate sufficiently long sequences while maintaining a high acceptance rate.
This results from its design objective, which jointly maximizes generation efficiency and the acceptance rate.
While AssistedGen occasionally adopts a conservative strategy that underestimates the candidate sequence length, it terminates generation at an appropriate point without excessive token generation, thus minimizing the draft model's inference overhead.

In contrast, AdaSD's generation threshold relies on the uncertainty of tokens produced by the draft model.
Therefore, this can lead to relatively more candidate tokens and the likelihood of tokens being rejected, which increases both generation and verification overhead.
This explains why AssistedGen outperforms AdaSD with Qwen.

We further investigate whether combining the strategies of AdaSD and AssistedGen can improve decoding efficiency for Qwen.
As shown in Table~\ref{tab:qwen_analysis}, the hybrid approach achieves the best performance across all tasks.
On GSM8K, the speedup increases from 1.287 (AssistedGen) and 1.257 (AdaSD) to 1.307 with the hybrid approach.
Similarly, the speedup improves from 1.282 and 1.250 to 1.286 on HumanEval, and from 1.088 and 1.091 to 1.134 on MMLU.
These results indicate that AdaSD’s adaptive verification and AssistedGen’s generation-length control mechanism are complementary, and their combination enables further improvements in speculative decoding efficiency.


\begin{table*}[t!]
    \centering
    \adjustbox{max width=\linewidth}{
        \begin{tabular}{c|rrrr|rrrr|rrrr}
    \hline
    \multicolumn{13}{c}{\textbf{Qwen 2.5 72B -- Qwen 2.5 7B}} \\
    \hline
    & \multicolumn{4}{c|}{\textbf{GSM8K}} & \multicolumn{4}{c|}{\textbf{HumanEval}} & \multicolumn{4}{c}{\textbf{MMLU}} \\
    \hline
    & \#cand & \#match & AccRate & speedup
    & \#cand & \#match & AccRate & speedup
    & \#cand & \#match & AccRate & speedup \\
    \hline
    Vanilla & 5.000 & 4.228 & 0.846 & 1.000 & 5.000 & 3.942 & 0.788 & 1.000 & 5.000 & 2.811 & 0.562 & 1.000 \\
    AssistedGen & 10.038 & 8.335 & 0.830 & \underline{1.287} & 9.745 & 7.720 & 0.792 & \underline{1.282} & 6.109 & 3.618 & 0.592 & 1.088 \\
    \hdashline
    Gen-Only & 14.254 & 9.967 & 0.699 & 1.241 & 13.199 & 8.832 & 0.669 & 1.227 & 8.962 & 4.210 & 0.470 & 1.021 \\
    Verify-Only & 5.000 & 4.290 & 0.863 & 1.011 & 5.000 & 4.024 & 0.812 & 1.015 & 5.000 & 3.010 & 0.610 & 1.051 \\
    AdaSD & 14.389 & 10.541 & 0.731 & 1.257 & 13.476 & 9.141 & 0.677 & 1.250 & 8.904 & 4.539 & 0.495 & \underline{1.091} \\
    \hdashline
    AdaSD + AssistedGen & 10.297 & 8.639 & 0.840 & \textbf{1.307} & 9.646 & 7.727 & 0.800 & \textbf{1.286} & 5.761 & 3.743 & 0.646 & \textbf{1.134} \\
    \hline
\end{tabular}
    }
    \caption{
        Decoding results of Qwen on different datasets.
        The result shows the number of generated candidates (\#cand), accepted tokens (\#match), and the acceptance rates (AccRate) collectively influence the decoding speedup.
    }
    \label{tab:qwen_analysis}
\end{table*}

\subsection{Ablation Studies}

In this section, we analyze the individual impact of AdaSD's components 
by comparing AdaSD with its two variants, \emph{Gen-Only}, which applies only the generation component, and \emph{Verify-Only}, which applies only the verification component, as reported in Tables~\ref{tab:main_result} and \ref{tab:qwen_analysis}.
For Verify-Only, we fix the candidate token length to five, consistent with Vanilla.

\paragraph{Gen-Only.}
Gen-Only demonstrates strong inference speedup over Vanilla, achieving a 16-42\% improvement on GSM8K and HumanEval.
On MMLU, the improvement is more modest, ranging from 1.4\% to 3.9\% due to the inherently short generation sequences.
This improvement arises from generating longer candidate token sequences.
The draft model continues generating tokens until the token entropy exceeds the generation threshold, which is adaptively updated based on feedback from the target model.

\paragraph{Verify-Only.}
In contrast, the speedup of Verify-Only is limited when used alone, with at most 6.9\% improvement over Vanilla on MMLU.
This is because Vanilla achieves a high acceptance rate with its carefully chosen candidate length, effectively covering most general cases.
However, when combined with the generation component, AdaSD further improves the efficiency, bringing the total improvement to 12\% over Vanilla (5.1\% over Verify-Only).
These results indicate that our JS distance-based acceptance criteria can further accelerate inference while maintaining accuracy.
\section{Conclusion}
This paper proposes AdaSD, a novel speculative decoding method without the need for manual hyperparameter tuning.
AdaSD improves decoding speed with minimal accuracy loss.
While AdaSD enhances decoding efficiency, there is still room for further improvement in its adaptive mechanisms.
For generation, incorporating additional indicators beyond entropy could enable more precise decisions.
For verification, refining JS distance modeling could further boost acceptance and accuracy.
We believe that AdaSD represents a practical step toward more efficient and adaptive decoding for large language models.

\section*{Limitations}
AdaSD is applicable when the draft and target models share the same vocabulary, meaning that both models must have identical token sizes and token content.
This requirement arises because AdaSD relies on computing entropy and JS distance for output tokens.
When the vocabulary differs, these metrics cannot be calculated consistently, and determining an appropriate threshold becomes difficult.

Another limitation lies in the challenge of selecting compatible draft and target model pairs.
In practice, the optimal size gap between the draft and target models is roughly 50-100 times.
However, within this range, few well-known model pairs share an identical vocabulary.
This issue is exacerbated by recent model development trends, where smaller models (<1B) often use different vocabularies from larger-sized models (>70B).
Moreover, smaller models frequently incorporate architectural modifications beyond simply reducing the number of Transformer layers, such as changes to embedding layers or tokenization schemes, which further limits compatibility and applicability of AdaSD.

\bibliography{references}

\appendix
\section{Information Theoretic Background}
\label{sec:information_theoretic_background}

\paragraph{Entropy.}
In information theory, entropy measures the uncertainty associated with a random variable.
For a discrete random variable $X$ taking values $x \in \mathcal{X}$ with a probability distribution $p: \mathcal{X} \rightarrow [0, 1]$, the entropy is defined as:
\begin{equation*}
    \begin{split}
        H(X) &= \mathbb{E}[-\log p(x)] \\
             &= -\sum_{x \in \mathcal{X}} p(x) \log p(x).
    \end{split}
\end{equation*}
Entropy can be interpreted as the expected value of the self-information $-\log p(x)$, representing the average number of bits needed to encode the outcome of $X$.  
A larger entropy value indicates higher uncertainty or greater variability in the distribution.

\paragraph{Cross-Entropy.}
Cross-entropy quantifies the difference between two probability distributions $p$ and $q$ defined over the same event space $\mathcal{X}$.
Formally, the cross-entropy of an estimated distribution $q$ with respect to a true distribution $p$ is defined as:
\begin{equation*}
    \begin{split}
        H(p, q) &= \mathbb{E}_p[-\log q(x)] \\
                &= -\sum_{x \in \mathcal{X}} p(x) \log q(x).
    \end{split}
\end{equation*}
Cross-entropy represents the expected number of bits needed to encode samples drawn from $p$ using a code optimized for $q$.  
Hence, it captures both the inherent uncertainty of $p$ and the additional cost due to the divergence between $p$ and $q$.

\paragraph{Kullback-Leibler Divergence.}
The Kullback-Leibler (KL) divergence measures the discrepancy between a true distribution $p$ and an approximating distribution $q$.
It can be expressed as the difference between cross-entropy and entropy:
\begin{equation*}
    \begin{split}
        D_\mathrm{KL}(p \parallel q) &= H(p, q) - H(p) \\
                                     &= \sum_{x \in \mathcal{X}} p(x) \log \frac{p(x)}{q(x)}.
    \end{split}
\end{equation*}
Since KL divergence has no fixed upper bound, if an event occurs frequently under $p$ but rarely under $q$, the divergence can become arbitrarily large, making it difficult to stop or normalize the measure within a controlled range.
\section{Method Comparison}
\label{sec:method_comparison}

This section provides a summary comparison of representative speculative decoding methods.
Table~\ref{tab:method_comparison} highlights several key design choices across prior work and our proposed method.
Specifically, we compare whether each method adopts self-speculative decoding, dynamically adjusts the draft sequence length, or introduces relaxed verification mechanisms.
The table also indicates whether the method requires architectural modifications or additional training, relies on empirically tuned hyperparameters, or performs task-specific pre-analysis.
This comparison helps clarify the design space of speculative decoding methods and highlights the characteristics of AdaSD.

\begin{table*}[t!]
    \centering
    \adjustbox{max width=\linewidth}{
        \begin{tabular}{c|c|cc|ccc}
\hline
paper & \makecell{self-\\speculative} & \makecell{dynamic\\draft\\length} & \makecell{relaxed\\verification} & \makecell{architecture\\modification/\\extra training} & \makecell{empirically tuned\\hyperparameter}& \makecell{task\\pre-analyzing}\\
\hline
Medusa~\cite{cai2024medusa}             & V &   &   & V &   &   \\
EAGLE~\cite{li2024eagle, li2024eagle2}  &   & V &   & V &   &   \\
\hdashline
Draft \& Verify~\cite{zhang2024draft}   & V & V &   &   & V & V \\
EESD~\cite{liu2024speculative}          & V & V &   & V & V &   \\
SWIFT~\cite{xia2025swift}               & V & V &   &   & V &   \\
\hdashline
DISCO~\cite{mamou2024dynamic}           &   & V &   & V &   & V \\
AdaEDL~\cite{agrawal2024adaedl}         &   & V &   &   & V &   \\
SpecDec++~\cite{huang2025specdec}       &   & V &   & V &   &   \\
PEARL~\cite{liu2025pearl}               &   & V &   &   &   & V \\
HSDDW~\cite{syu2025hierarchical}        &   & V &   &   &   &   \\
C2T~\cite{huo2026c2t}                   &   & V &   & V & V &   \\
\hdashline
BiLD~\cite{kim2023speculative}          &   & V & V & V &   & V \\
FSD~\cite{holsman2025fuzzy}             &   &   & V &   & V & V \\
\hdashline
AssistedGen~\cite{gante2023assisted}    &   & V &   &   &   &   \\
AdaSD~(Ours)                            &   & V & V &   &   &   \\
\hline
\end{tabular}
    }
    \caption{Comparison of representative speculative decoding methods and their key design choices.}
    \label{tab:method_comparison}
\end{table*}

\section{Algorithm of AdaSD}
\label{sec:algorithm_of_adasd}

Algorithm~\ref{algo:adasd} shows the AdaSD procedure, which comprises three main steps: generation, verification, and update.
To enable adaptive threshold adjustment, we record the entropy of generated tokens and the JS distances during inference in three lists: rejected entropy ($L_E^R$), rejected JS distance ($L_D^R$), and accepted JS distance ($L_D^A$).
These statistics are used to update the generation and verification thresholds.
Moreover, two additional constraints are imposed.
First, we set the maximum window size $W$ to 20, limiting the number of candidate tokens generated per iteration and preventing unbounded draft expansion.
Second, due to the inherent maximum context length of LLMs, we define a limit $K$ on the total number of generated tokens to ensure the context $X$ remains within allowable bounds.

\begin{algorithm*}[t!]
    \caption{Adaptive Speculative Decoding}
    \label{algo:adasd}
    \SetKwInOut{Initialization}{Initialization}
    \SetKw{Break}{break}
    \SetKwFunction{Append}{append}
    \SetKwFunction{Average}{average}
    \SetKwFunction{KDE}{KDE}
    
    \KwIn{target model $M_p$, draft model $M_q$, context $X$, max tokens $K$, max window size $W$}
    \KwOut{final generated context $X$}
    
\Initialization{
    \begin{tabular}[t]{ll}
        $L_E^R \gets [\,]$ & \tcp*{\textbf{L}ist of \textbf{E}ntropies (\textbf{R}ejected)} \\
        $L_D^A, L_D^R \gets [\,], [\,]$ & \tcp*{\textbf{L}ists of JS \textbf{D}istances (\textbf{A}ccepted/\textbf{R}ejected)} \\
        $T_G \gets 0$ & \tcp*{\textbf{G}eneration \textbf{T}hreshold} \\
        $w \gets 0$ & \tcp*{\textbf{W}indow size counter}
    \end{tabular}
}
    
    \While{$len(X) < K$}{
        \tcp{1. The draft model generates tokens sequentially}
        \For{$i \gets 1$ \KwTo $W$}{
            $q_i \gets M_q(X, x_{1:i-1})$ \tcp*{Define $x_{1:0}$ is null}
            $x_i \sim q_i$ \tcp*{Sample $x_i$ from $q_i$}
            $w \gets i$ \tcp*{Record generated window length}
            \If(\tcp*[f]{Check generation threshold}){$H(q_i) > T_G$}{
                \Break\;
            }
        }
        
        \tcp{2. The target model verifies tokens in parallel}
        $p_1, \dots, p_{w+1} \gets M_p(X, x_{1:0}), \dots, M_p(X, x_{1:w})$\;
        $y_1, \dots, y_{w+1} \sim p_1, \dots, p_{w+1}$ \tcp*{sample $y_i$ from $p_i$}
        
        \For{$i \gets 1$ \KwTo $w$}{
            $d_i \gets d_{\mathrm{JS}}(p_i \parallel q_i)$\;
            $p(d_i \mid A) \gets \KDE(L_D^A, d_i)$ \tcp*{density under accepted distances}
            $p(d_i \mid R) \gets \KDE(L_D^R, d_i)$ \tcp*{density under rejected distances}
            
            $P(A) \gets \frac{|L_D^A|}{|L_D^A|+|L_D^R|}$ \tcp*{prior of acceptance}
            $P(R) \gets \frac{|L_D^R|}{|L_D^A|+|L_D^R|}$ \tcp*{prior of rejection}
            
            $P(A \mid d_i) \gets 
            \frac{p(d_i \mid A)\,P(A)}
            {p(d_i \mid A)\,P(A) + p(d_i \mid R)\,P(R)}$ \tcp*{acceptance probability}
            
            \If{$x_i \neq y_i$ \textbf{and} \textnormal{Reject}($P(A \mid d_i)$)}{
                $L_E^R.\Append(H(q_i))$ \tcp*{record rejected entropy}
                \If{$0 < d_i < 1$}{
                    $L_D^R.\Append(d_i)$ \tcp*{record rejected JS distance}
                }
                $w \gets i - 1$ \tcp*{record accepted window length}
                \Break\;
            }
            \If{$0 < d_i < 1$}{
                $L_D^A.\Append(d_i)$ \tcp*{record accepted JS distance}
            }
        }

        \tcp{3. Update parameters}
        $X.\Append(x_{1:w}, y_{w+1})$ \tcp*{update context}
        $T_G \gets \Average(L_E^R)$ \tcp*{update generation threshold}
    }
\end{algorithm*}
\section{Implementation Details}
\label{sec:implementation_details}

Experiments are implemented using the Hugging Face Transformer framework~\cite{wolf2020transformers} version 4.55.
The Transformer backend uses PyTorch~\cite{paszke2019pytorch} version 2.8 with CUDA 12.8 and cuDNN 9.10.
We run all tests on four NVIDIA A6000 GPUs and enable ``\texttt{device\_map=auto}'' to evenly split the model on all available GPUs.
All evaluated methods use speculative sampling for token selection.

\section{Additional Experimental Results}
\label{sec:additional_experimental_results}

Tables~\ref{tab:gsm8k}, \ref{tab:humaneval}, and~\ref{tab:mmlu} present the performance results on the GSM8K, HumanEval, and MMLU datasets respectively.

\begin{table*}[t!]
    \centering
    \adjustbox{max width=.6\linewidth}{
        \begin{tabular}{c|rrrrrr}
    \hline
    \multicolumn{7}{c}{\textbf{openai/gsm8k, main, test, num\_rows = 1319}} \\
    \hline
    \hline
    \multicolumn{7}{c}{\textbf{Llama 3.1 70B -- Llama 3.1 8B}} \\
    \hline
    & tks/sec & \#cand & \#match & AccRate & accuracy & speedup \\
    \hline
    Draft & 36.446 & - & - & - & 0.828 & 2.624 \\
    Target & 4.842 & - & - & - & 0.939 & 0.349 \\
    Vanilla & 13.890 & 5.000 & 4.057 & 0.811 & 0.945 & 1.000 \\
    AssistedGen & 16.062 & 7.898 & 6.311 & 0.799 & 0.944 & 1.156 \\
    Gen-Only & 16.243 & 11.157 & 7.850 & 0.704 & 0.943 & 1.169 \\
    Verify-Only & 14.227 & 4.969 & 4.189 & 0.843 & 0.931 & 1.024 \\
    AdaSD & 17.039 & 11.168 & 8.309 & 0.745 & 0.943 & 1.227 \\
    \hline
    \hline
    \multicolumn{7}{c}{\textbf{Llama 3.1 70B -- Llama 3.2 1B}} \\
    \hline
    & tks/sec & \#cand & \#match & AccRate & accuracy & speedup \\
    \hline
    Draft & 99.919 & - & - & - & 0.375 & 6.080 \\
    Target & 4.842 & - & - & - & 0.939 & 0.295 \\
    Vanilla & 16.434 & 5.000 & 3.596 & 0.719 & 0.939 & 1.000 \\
    AssistedGen & 19.892 & 6.918 & 5.047 & 0.730 & 0.936 & 1.210 \\
    Gen-Only & 21.126 & 10.438 & 6.189 & 0.593 & 0.939 & 1.286 \\
    Verify-Only & 17.063 & 4.967 & 3.748 & 0.755 & 0.926 & 1.038 \\
    AdaSD & 22.225 & 10.358 & 6.560 & 0.636 & 0.937 & 1.352 \\
    \hline
    \hline
    \multicolumn{7}{c}{\textbf{Qwen 2.5 72B -- Qwen 2.5 7B}} \\
    \hline
    & tks/sec & \#cand & \#match & AccRate & accuracy & speedup \\
    \hline
    Draft & 38.796 & - & - & - & 0.867 & 2.667 \\
    Target & 4.728 & - & - & - & 0.913 & 0.325 \\
    Vanilla & 14.546 & 5.000 & 4.228 & 0.846 & 0.912 & 1.000 \\
    AssistedGen & 18.721 & 10.038 & 8.335 & 0.830 & 0.917 & 1.287 \\
    Gen-Only & 18.054 & 14.254 & 9.967 & 0.699 & 0.920 & 1.241 \\
    Verify-Only & 14.705 & 4.971 & 4.290 & 0.863 & 0.920 & 1.011 \\
    AdaSD & 18.289 & 14.389 & 10.514 & 0.731 & 0.915 & 1.257 \\
    \hline
\end{tabular}
    }
    \caption{Inference results on the GSM8K test set with different schemes across three model pair combinations.}
    \label{tab:gsm8k}
\end{table*}

\begin{table*}[t!]
    \centering
    \adjustbox{max width=.6\linewidth}{
        \begin{tabular}{c|rrrrrr}
    \hline
    \multicolumn{7}{c}{\textbf{openai/openai\_humaneval, test, num\_rows = 164}} \\
    \hline
    \hline
    \multicolumn{7}{c}{\textbf{Llama 3.1 70B -- Llama 3.1 8B}} \\
    \hline
    & tks/sec & \#cand & \#match & AccRate & accuracy & speedup \\
    \hline
    Draft & 36.257 & - & - & - & 0.634 & 2.532 \\
    Target & 4.838 & - & - & - & 0.750 & 0.338 \\
    Vanilla & 14.319 & 5.000 & 4.264 & 0.853 & 0.768 & 1.000 \\
    AssistedGen & 18.137 & 9.095 & 7.927 & 0.872 & 0.756 & 1.267 \\
    Gen-Only & 18.017 & 13.274 & 9.972 & 0.751 & 0.762 & 1.258 \\
    Verify-Only & 14.425 & 4.968 & 4.318 & 0.869 & 0.750 & 1.007 \\
    AdaSD & 18.348 & 13.015 & 10.054 & 0.772 & 0.799 & 1.281 \\
    \hline
    \hline
    \multicolumn{7}{c}{\textbf{Llama 3.1 70B -- Llama 3.2 1B}} \\
    \hline
    & tks/sec & \#cand & \#match & AccRate & accuracy & speedup \\
    \hline
    Draft & 100.683 & - & - & - & 0.348 & 5.764 \\
    Target & 4.838 & - & - & - & 0.750 & 0.277 \\
    Vanilla & 17.467 & 5.000 & 3.915 & 0.783 & 0.756 & 1.000 \\
    AssistedGen & 23.265 & 7.759 & 6.362 & 0.820 & 0.774 & 1.332 \\
    Gen-Only & 24.842 & 11.962 & 7.930 & 0.663 & 0.768 & 1.422 \\
    Verify-Only & 17.710 & 4.969 & 3.961 & 0.797 & 0.726 & 1.014 \\
    AdaSD & 25.452 & 11.829 & 8.124 & 0.686 & 0.774 & 1.457 \\
    \hline
    \hline
    \multicolumn{7}{c}{\textbf{Qwen 2.5 72B -- Qwen 2.5 7B}} \\
    \hline
    & tks/sec & \#cand & \#match & AccRate & accuracy & speedup \\
    \hline
    Draft & 38.724 & - & - & - & 0.720 & 2.840 \\
    Target & 4.717 & - & - & - & 0.787 & 0.346 \\
    Vanilla & 13.634 & 5.000 & 3.942 & 0.788 & 0.799 & 1.000 \\
    AssistedGen & 17.482 & 9.745 & 7.720 & 0.792 & 0.799 & 1.282 \\
    Gen-Only & 16.733 & 13.199 & 8.832 & 0.669 & 0.787 & 1.227 \\
    Verify-Only & 13.838 & 4.956 & 4.024 & 0.812 & 0.787 & 1.015 \\
    AdaSD & 17.041 & 13.476 & 9.141 & 0.677 & 0.799 & 1.250 \\
    \hline
\end{tabular}
    }
    \caption{Inference results on the HumanEval test set with different schemes across three model pair combinations.}
    \label{tab:humaneval}
\end{table*}

\begin{table*}[t!]
    \centering
    \adjustbox{max width=.6\linewidth}{
        \begin{tabular}{c|rrrrrr}
    \hline
    \multicolumn{7}{c}{\textbf{cais/mmlu, all, validation, num\_rows = 1531}} \\
    \hline
    \hline
    \multicolumn{7}{c}{\textbf{Llama 3.1 70B -- Llama 3.1 8B}} \\
    \hline
    & tks/sec & \#cand & \#match & AccRate & accuracy & speedup \\
    \hline
    Draft & 33.274 & - & - & - & 0.695 & 4.206 \\
    Target & 4.376 & - & - & - & 0.836 & 0.553 \\
    Vanilla & 7.912 & 5.000 & 2.162 & 0.432 & 0.835 & 1.000 \\
    AssistedGen & 7.970 & 5.581 & 2.482 & 0.445 & 0.835 & 1.007 \\
    Gen-Only & 8.023 & 6.297 & 2.726 & 0.433 & 0.831 & 1.014 \\
    Verify-Only & 8.247 & 4.402 & 2.285 & 0.511 & 0.836 & 1.042 \\
    AdaSD & 8.477 & 6.132 & 2.897 & 0.490 & 0.834 & 1.071 \\
    \hline
    \hline
    \multicolumn{7}{c}{\textbf{Llama 3.1 70B -- Llama 3.2 1B}} \\
    \hline
    & tks/sec & \#cand & \#match & AccRate & accuracy & speedup \\
    \hline
    Draft & 95.141 & - & - & - & 0.453 & 10.878 \\
    Target & 4.376 & - & - & - & 0.836 & 0.500 \\
    Vanilla & 8.746 & 5.000 & 1.714 & 0.343 & 0.838 & 1.000 \\
    AssistedGen & 8.920 & 4.759 & 1.845 & 0.388 & 0.833 & 1.020 \\
    Gen-Only & 9.085 & 5.205 & 1.956 & 0.376 & 0.835 & 1.039 \\
    Verify-Only & 9.346 & 4.226 & 1.903 & 0.438 & 0.817 & 1.069 \\
    AdaSD & 9.795 & 5.297 & 2.193 & 0.422 & 0.815 & 1.120 \\
    \hline
    \hline
    \multicolumn{7}{c}{\textbf{Qwen 2.5 72B -- Qwen 2.5 7B}} \\
    \hline
    & tks/sec & \#cand & \#match & AccRate & accuracy & speedup \\
    \hline
    Draft & 38.544 & - & - & - & 0.725 & 3.667 \\
    Target & 4.690 & - & - & - & 0.845 & 0.446 \\
    Vanilla & 10.511 & 5.000 & 2.811 & 0.562 & 0.837 & 1.000 \\
    AssistedGen & 11.436 & 6.109 & 3.618 & 0.592 & 0.837 & 1.088 \\
    Gen-Only & 10.735 & 8.962 & 4.210 & 0.470 & 0.841 & 1.021 \\
    Verify-Only & 11.042 & 4.920 & 3.010 & 0.610 & 0.837 & 1.051 \\
    AdaSD & 11.470 & 8.904 & 4.539 & 0.495 & 0.839 & 1.091 \\
    \hline
\end{tabular}
    }
    \caption{Inference results on the MMLU validation set with different schemes across three model pair combinations.}
    \label{tab:mmlu}
\end{table*}


\end{document}